\documentclass[conference]{IEEEtran}

\usepackage[pdftex]{graphicx}
\usepackage{hyperref}
\usepackage{subcaption}

\title{Deep Architectures for Modulation Recognition}

\author{\IEEEauthorblockN{Nathan E. West}
\IEEEauthorblockA{U.S. Naval Research Laboratory\\
Washington, D.C\\
Email: nathan.west@nrl.navy.mil}
\and
\IEEEauthorblockN{Timothy J. O'Shea}
\IEEEauthorblockA{Virginia Polytechnic Institute and State University\\
Arlington, VA\\
Email: oshea@vt.edu}
}

\graphicspath{{../sparse-geo/}}

\begin{document}
\maketitle

\begin{abstract}

We survey the latest advances in machine learning with deep neural networks by
applying them to the task of radio modulation recognition. Results show that
radio modulation recognition is not limited by network depth and further work
should focus on improving learned synchronization and equalization. Advances in
these areas will likely come from novel architectures designed for these tasks
or through novel training methods.

\end{abstract}

\section{Introduction}

Deep neural networks have been pushing recent performance boundaries for a
variety of machine learning tasks in fields such as computer vision, natural
language processing, and speaker recognition. Recently researchers in the
wireless communications field have started to apply deep neural networks to
cognitive radio tasks with some success \cite{conv-modrec, grcon,
bio-inspired-modrec}. In particular it has been shown that relatively simple
convolutional neural networks outperform algorithms with decades of expert
feature searches for radio modulation \cite{conv-modrec}. This paper provides
an introduction to deep neural networks for the cognitive radio task of
modulation recognition, compares several state of the art methods in other
domains, and experiments with learning techniques.

Deep neural networks are large function approximations comprised of a series of
layers, where each layer represents some transform from input to output activations
based on a parametric transfer function with some set of learned weights.
Each layer is typically a known linear function with adjustable parameters and a non-linear
activation function such that the resulting function composition can be highly
non-linear \cite{dl-2016-book}.  Function parameters in deep neural networks
are typically trained with a gradient descent optimizer from some loss
function computed on the output of the network. For a multi-class
classification task such as modulation recognition the objective function is
often categorical cross-entropy (eq. \ref{eq:cat-cross-entropy}). Categorical
cross-entropy is a measure of difference between two probability distributions.
For deep learning classification tasks the probability distribution is usually
a softmax (eq. \ref{eq:softmax}) of the output of the classifier network which
is then converted to a one-hot encoding for classification purposes
\cite{dl-2016-book}.  The error is calculated in what is known as the
forward-pass and weights are adjusted using the chain rule to find error
contribution for each parameter in what is known as the backward-pass.
This kind of network output layer, optimization and loss function have been
used very successfully for multi-class vision tasks such as object recognition
on the Imagenet dataset \cite{krizhevsky2012imagenet}.

\begin{equation}
    \label{eq:cat-cross-entropy}
    H(p,q) = \Sigma_x p(x) \log q(x)
\end{equation}

\begin{equation}
    \label{eq:softmax}
    \sigma(z)_j = \frac{e^{z_j}}{\Sigma^K _{k=1} e^{z_k}} \quad \mathrm{for}\  j=1,...,K
\end{equation}

Applying deep neural networks to solve well-known problem types, such as
classification, is a matter of
\begin{itemize}
  \item selecting a network architecture and hyper-parameters
  \item training the network to select weights which minimize loss
  \item applying it to the problem at hand
\end{itemize}

There are several well established network architectures including
multi-layer perceptrons, many variations of convolutional networks, and recurrent
networks. Although the goal of machine learning is to develop generalized
techniques the current state of the art network types still seem to be
application specific. For example, Google views Convolutional Long short-term
Deep Neural Networks (CLDNN) to be worthy of patenting even though it is only
used in their voice processing research. The state of the art in image
recognition uses variants of the inception architecture, residual networks, and
other architectures that enable many combinations of convolutional layers, while
managing the combinatorial complexity of weights and activations.  We discuss these
methods in greater detail in the next section.

Before applying deep neural networks to wireless communications signals it is worth
reviewing the state of the art for other application areas. The next section
will review deep neural network architecture and learning advances that are
likely to be valid and useful for wireless communications applications.
Following the review of interesting deep architectures and training methods,
results are in section \ref{sec:results} and a discussion in section
\ref{sec:discussion}.

\subsection{Neural Network Architectures}

The common element in all state of the art deep neural networks is the use of
convolutional layers. A convolutional layer consists of $N_f$ convolutional
filters. The use of convolutional layers started for image and hand-writing
recognition to provide feature translation invariance \cite{lecun-convnets}. The
use of convolutional filters in neural networks may be slightly different than
expected for someone already familiar with FIR filters and DSP at least
partially due to the use of activation functions in neural networks.
Convolutions in neural networks are typically very small (1x1 through 5x5 are
common sizes in image processing). In typical DSP applications filters are very
wide (many taps/high order) rather than deep (small taps, but cascaded). Modern
methods of implementing these filters, such as polyphase filterbanks, typically
provide ways to reduce the width of filters for computational or latency
reasons.  The transfer function for a standard convolutional layer \cite{cnnppr}
is given below in equation \ref{eqn:cnn}, where $y_i$ is the output feature map
for the $i$th filter, $b$ and $k$ represent learned bias and filter weight
parameters, $x_i$ represents the input activations, $\ast$ denotes the
convolution operation, and $f(..)$ denotes a (typically non-linear) activation
function such as a rectified linear unit (ReLU) or sigmoid.

\begin{equation}
    \label{eqn:cnn}
    y_i = f\left( b_j + \Sigma_i k_{i j} \ast x_i \right)
\end{equation}

A visible trend in neural networks for image processing is building deeper
networks to learn more complex functions and hierarchical feature relationships
\cite{understanding-deep-archs, deepnets-deep}. Deep networks enable more
complex functions to be learned more readily from raw data than shallower
networks with the same number of parameters \cite{deepnets-deep,
deepconvnets-deepconv}; however, depth in neural networks is widely believed to
be limited by unstable gradients that either explode or vanish in earlier or
later layers in the network. This problem has been improved in recent years by
the use of gradient normalization in optimizers as well as non-linearities which
do not exacerbate the vanishing gradient problem such as rectified linear units (ReLUs).
As a result several important architectures have
been used to win competitions such as ImageNet by increasing depth that we will
look at for improving radio modulation recognition.

\begin{figure}
  \includegraphics[width=\columnwidth]{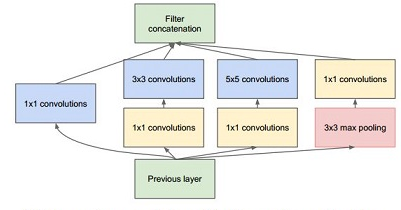}
  \caption{\label{fig:inception} Inception Unit diagram from \cite{inception-arch}, generalizing to feature learning of multiple scales while also managing model complexity.}
\end{figure}

The inception architecture used in GoogLenet \cite{inception-arch} is one
successful approach to increasing network depth and ability to generalize to
feature of differing scales while still managing complexity. This network consists of
repeated inception modules. Each inception module (shown in figure \ref{fig:inception}),
contains four parallel paths
with the output being the concatenation of the four parallel outputs. The first
path is a bank of 1x1 convolutions that forward along selected information. The
1x1 convolutions are a form of selective highway networks that simply pass
information forward with no transformation. The second and third paths are 1x1
convolutions followed by a bank of 3x3 and 5x5 convolutions to provide multiple
scales of feature detection. Finally, the last parallel path is a 3x3 pooling
layer followed by 1x1 convolutions. Intermediate inception modules in the
network are connected to softmax classifiers that contribute to the network's
global loss for training.  These classifiers are believed to help in preventing
vanishing gradients.

\begin{figure}
  \includegraphics[width=\columnwidth]{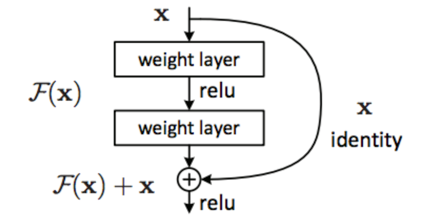}
  \caption{\label{fig:resnet} Residual network diagram from \cite{deep-resnets},
  allowing feature maps combinations from multiple source layers to select optimal
  architecture paths within a network.}
\end{figure}

Another approach to increasing depth uses architectures that forward
information untouched across layers. The best approach so far, which won
ImageNet 2015, is residual networks \cite{deep-resnets}. A residual network
adds one layer's output to the output of the layer two layers deeper (as shown
in figure \ref{fig:resnet}).  This is
known as a residual network because the forwarded information forces the
network to learn a residual function as part of feature extraction. The
residual network authors suggest that vanishing gradients are resolved by
normalization techniques that have been widely adopted and that network depth
is instead limited by training complexity of deep networks which can be
simplified with residual functions.

CLDNNs are an approach for voice processing that operate on raw time-domain
waveforms rather than expert voice features such as log-mel cepstrums
\cite{raw-acoustic-cldnn, cldnn}.  The architecture uses two convolutional
layers followed by two recurrent layers made up of Long Short-Term Memory
(LSTM) cells. LSTMs are a common recurrent network architecture consisting of
several gates that control how long history is maintained \cite{lstm}. CLDNNs
can also have connections that bypass layers that are intended to provide a
longer time context for the extracted features. For example, the original CLDNN
forwards raw samples with the output of convolutional layers before the LSTM
layers \cite{cldnn}.

Inspired by the use of expert knowledge to guide network architectures such as
convolutional networks and CLDNNs we experiment with a convolutional network
that we will refer to as a convolutional matched filter. The rather simple idea
is to take the general architecture of a typical communications receiver and
build a neural network architecture that has similar parts. Communications
receivers have a filter (typically matched to the transmitted pulse or wave
shape), synchronizer, and sampler. Often the filter up front decimates to a
small number of samples per symbol for the synchronizer which performs phase
shifts to find the optimal sampling point. The sampler then slices to bits or
emits audio for analog modulations. The neural network architecture analog to
this is a convolutional layer with pooling followed by an LSTM.

\subsection{Neural Network Training}

Hyper-parameters of a network such as learning rate, number of filters/feature
maps per layer, filter size, and to some extent number of layers all affect
network size and are hard to optimize. Recent research has attempted to
optimize hyper-parameters as regular parameters that can be trained with
backpropagation and gradient descent like network weights and biases. For this
study we ignore training hyper-parameters and use the adam optimizer
\cite{adam-optimizer} which provides gradient normalization and momentum which
reduces the importance of hyper-parameters like learning rate.

Guided by work that shows depth being more important than number of feature
maps \cite{understanding-deep-archs} we will establish a baseline convolutional
network similar to that used in Radio Convolutional Modulation Networks
\cite{conv-modrec}. Our first step is to tune the number of filters and number
of taps per filter and view those as unimportant hyper-parameters for the
remainder of experiments to test suitability of different architectures for RF
data.

\subsection{Test Setup}

\section{Technical Approach}

We use the RadioML2016.10a dataset \cite{grcon} as a basis for evaluating the modulation
recognition task. The goal is to use a 128-sample complex (baseband I/Q) time-domain
vector to identify the modulation scheme out of 11 possible classes. The 128
samples are fed in to the network in a 2x128 vector where the real and imaginary
parts of the complex time samples are separated. The dataset uses a power delay
profile, frequency selective fading, local oscillator offset, and additive white
Gaussian noise with details of these effects in \cite{grcon}. The dataset
is labeled with both modulation type and SNR ground truth.  We use the all-SNR top-1
classification accuracy as a single-number benchmark and show top-1 accuracy
over SNR to compare techniques.

All models and training are done with the Keras deep learning library using the
theano backend using an Nvidia GTX 1070 GPU.

We start with a network similar to the CNN2 network from
\cite{conv-modrec}. This is the chosen baseline because results from
\cite{conv-modrec} show significant improvement upon expert methods; any
further improvements should be considered state of the art. The primary
difference is we will use $nfilt$ filters of size 1x$taps$ on each layer. We
will do a simple hyper-parameter optimization to

\begin{itemize}
  \item find the best number of filters and filter size for RF modulation recognition
  \item test assumptions gained from other fields on network depth and filter size
\end{itemize}

\section{Results}
\label{sec:results}

\subsection{Baseline Convolution Network}

The baseline convolutional network has two convolution layers and a single
dense layer before the softmax classifier. Each hidden layer has a rectified
linear unit (ReLU) activation function and dropout of 50\%. The first hyper
parameter optimization is the size of our convolutional layers. Each layer will
have 1x3 filters and we will vary the number of filters to find how many are
needed. From
\cite{deepnets-deep,deepconvnets-deepconv,understanding-deep-archs} we expect
that a large range in the number of filters will give similar performance
before any overfitting will happen.

\begin{figure}
  \includegraphics[width=\columnwidth]{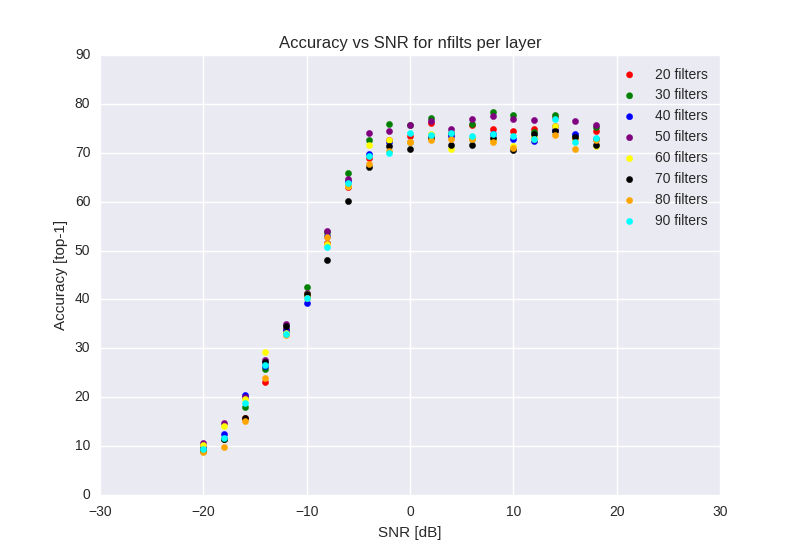}
  \caption{\label{fig:nfilts-hyperopt}Varying the number of filters per layer
has small impact that is more noticeable at higher SNRs. Each network has 1x3
filters in a 2 convolutional layer network with 1 dense layer and a softmax
classifier.}
\end{figure}

As expected there is a rather large window from about 30 to 70 filters per
layer where performance is very similar. The top-1 classification accuracy for
20-90 filters in 10-filter increments is shown in figure \ref{fig:nfilts-hyperopt}.
For the remaining experiments we will use 50 filters per layer.

Next, we optimize the size of each filter. \cite{understanding-deep-archs}
suggests that the size of filters also has minimal impact, but based on expert
knowledge of the radio domain and the dataset we expect 8-tap filters to be
optimal. For this experiment we use a two-convolution layer network with a single
hidden dense layer followed by the softmax classifier. The convolution layers each
have 50 filters with a filter size of 1x$ntaps$ where $ntaps$ varies from 3 to 12.

\begin{figure}
  \includegraphics[width=\columnwidth]{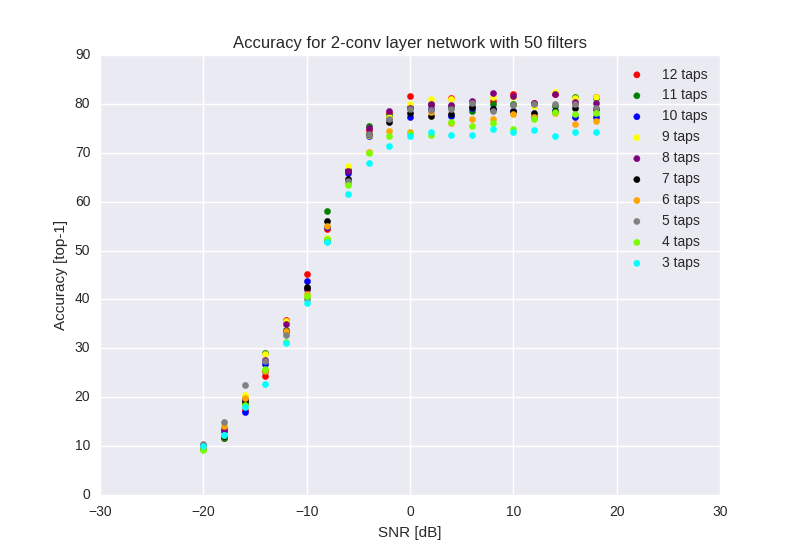}
  \caption{\label{fig:ntaps-hyperopt}Varying the number of taps (filter size)
in both convolutional layers. Lower numbers of taps are clearly inferior, but
performance clusters as number of taps increases.}
\end{figure}

Results from varying filter sizes for each convolution layer suggest that
smaller filters are not as good as larger filters. We hypothesized based on
expert knowledge of the dataset that 8-tap filters would be the best. It is
difficult to distinguish a clear winner from the results per SNR graph in figure
\ref{fig:ntaps-hyperopt}; however, the whole dataset classification accuracy
shows that 7-12 taps all have similar performance around 61\% with differences
being statistically insignificant.

Finally, for purely convolutional networks we experiment with increasing
network depth. For this experiment we use 50-tap convolutional layers with 1x8
filters. After the convolution layers we use a single hidden dense layer
followed by a final dense softmax classifier. We start with a
2-convolutional-layer network and add convolution layers. Trends from deep learning
suggest that adding more layers should improve classification performance until
the gradient becomes unstable.

\begin{figure}
  \includegraphics[width=\columnwidth]{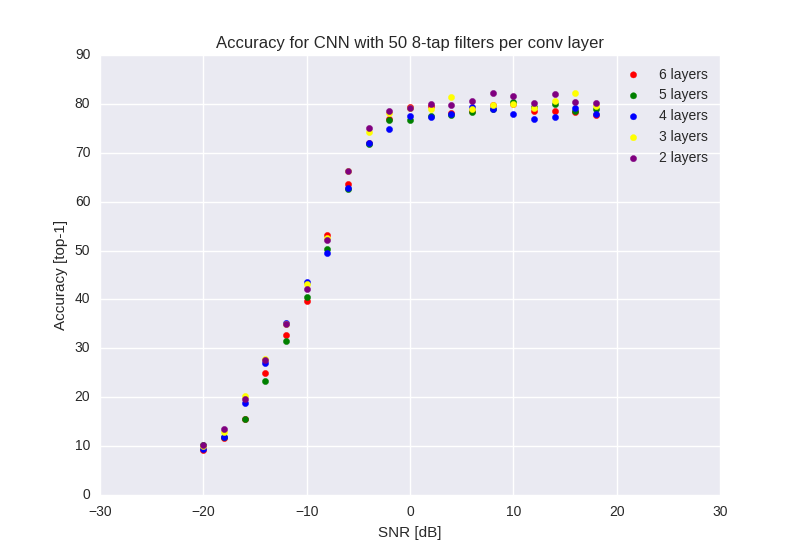}
  \caption{\label{fig:layers-hyperopt}Varying the number of convolution layers
in a DNN does not improve radio modulation recognition.}
\end{figure}

Varying the number of convolutional layers shows little to no improvement in
classification accuracy. Accuracy over SNR for this task is shown in figure
\ref{fig:layers-hyperopt}. This shows that there is no more feature depth for
our network to learn. The data is not highly hierarchical to start with since
the modulated data generally changes only the amplitude, frequency, or phase of
a complex sinusoid; however, it is somewhat surprising that adding more
convolutional layers does not appear to help reduce affects of noise at lower
SNRs.

\subsection{Residual Networks}

\begin{figure}
  \includegraphics[width=\columnwidth]{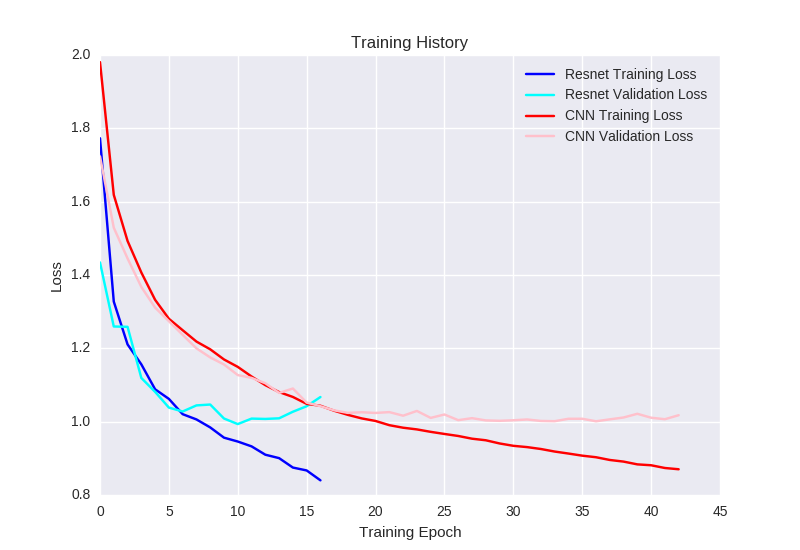}
  \caption{\label{fig:resnet-history}Training history showing the training loss
and validation loss for a hyper-parameter optimized CNN and a 9-layer residual
network. Both networks result in similar validation losses and training losses,
but the residual network trains in fewer epochs.}
\end{figure}

Although it is not surprising that adding more convolutional layers does not
improve classification accuracy it is surprising that the classification and
loss improvements plateau as soon as 2 or 3 convolutional layers. The original
resnet insight is that deeper networks result in higher training loss which
suggests higher training difficult rather than overfitting. Figure
\ref{fig:resnet-history} shows that our hyper-parameter optimized CNN and a
9-layer residual network reaches similar loss, validation loss, and accuracy
which is not shown; however, the residual network learns in fewer epochs. We
also experimented with residual networks with 5-9 layers that all had similar
performance and training times. This combined with our hyper-parameter search
for ordinary CNN depth suggests we are not limited by network depth for radio
learning tasks as much as we are limited by features purely CNN architectures
can learn.

\subsection{Inception Modules}

\begin{figure}
  \includegraphics[width=\columnwidth]{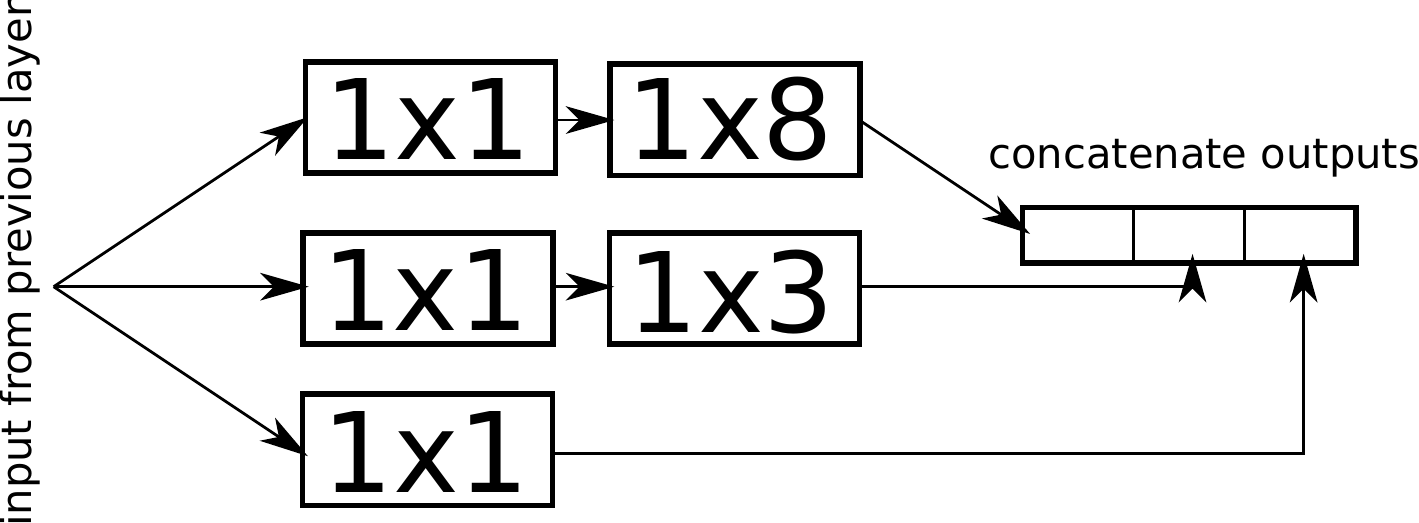}
  \caption{\label{fig:rf-inception}An inception module for RF data showing
convolution filter sizes. Each convolution layer has 50 filters.}
\end{figure}

Inception modules also do not improve radio modulation classification in our
experiments here, using inception modules tuned for our dataset.
The three branches used in each module are
50 1x1 filters, 50 1x3 filters, and 50 1x8 filters. The 1x3 and 1x5 filter
branches also have 50 1x1 filters in front of them as shown in figure
\ref{fig:rf-inception}. The results for 1-4 inception modules in a network do
not show any improvement over our hyper-parameter optimized CNN. Again, this
suggests that we are not limited by depth or apparently by scale of filters.

\subsection{LSTM Networks}

\begin{figure}
  \includegraphics[width=\columnwidth]{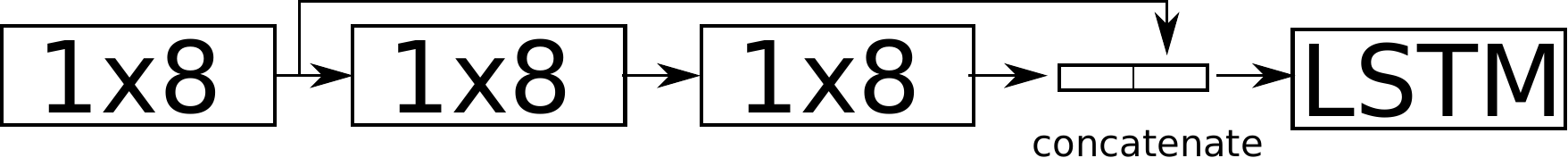}
  \caption{\label{fig:rf-cldnn}A CLDNN architecture for RF data. The output of
the first 1x8 convolution layer is concatenated with the output of three 1x8
convolution layers before going to the LSTM.}
\end{figure}

As the final architecture we test adding recurrent network layers, namely those
comprised of LSTM units, for modeling temporal features.
This approach is widely used in time-series applications and we expected that
modulated baseband time-series may be similarly applicable.
We tested two and three layer convolutions followed by recurrent layers in a
CLDNN-type architecture with and without the forward/bypass connection before the
recurrent layer. We found that the forward connection as a concatenation of the raw
waveform and the convolutional output, shown in figure \ref{fig:rf-cldnn}, results
in better classification accuracy and more stable gradient descent than other
architectures. Using a pooling layer that would create an architecture like the
previously described convolutional matched filter detector does not help
classification.

\begin{figure}
  \includegraphics[width=\columnwidth]{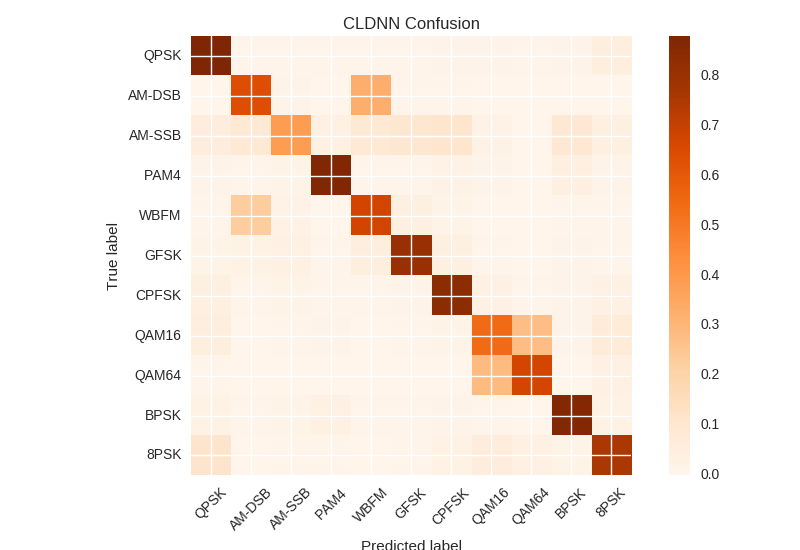}
  \caption{\label{fig:cldnn-confusion}The all-SNR confusion matrix for a CLDNN
shows the most confusion between analog modulations and a separate confusion
between higher order QAMs.}
\end{figure}

To further understand what limits classification accuracy we look at the
confusion matrix for a CLDNN shown in figure \ref{fig:cldnn-confusion}. There are
two primary areas of confusion. One is between the analog modulations and the
other is between higher order QAMs. The analog modulations will be hard to
address, but the QAMs can likely be improved on with better synchronization and
reducing channel impairments.

\begin{figure}
  \includegraphics[width=\columnwidth]{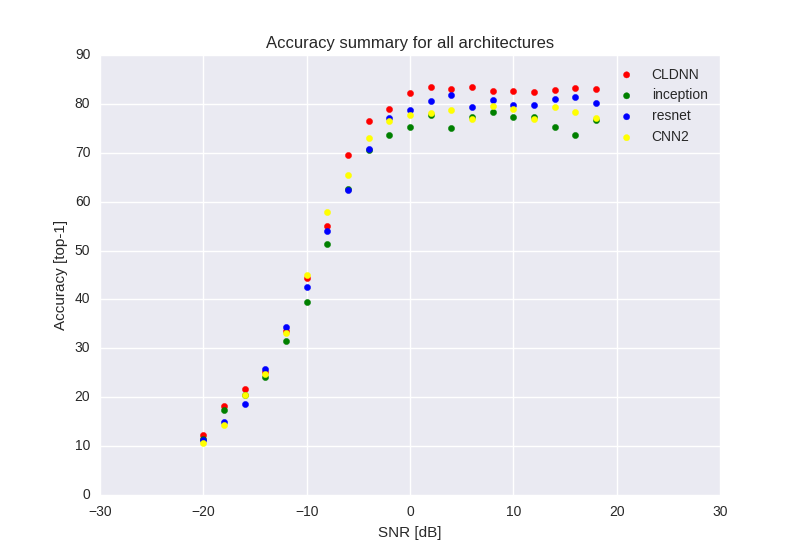}
  \caption{\label{fig:accuracy-summary}A CLDNN consistently outperforms
other network architectures for SNRs above -8dB.}
\end{figure}

\begin{figure}
  \begin{subfigure}{\columnwidth}
    \includegraphics[width=\columnwidth]{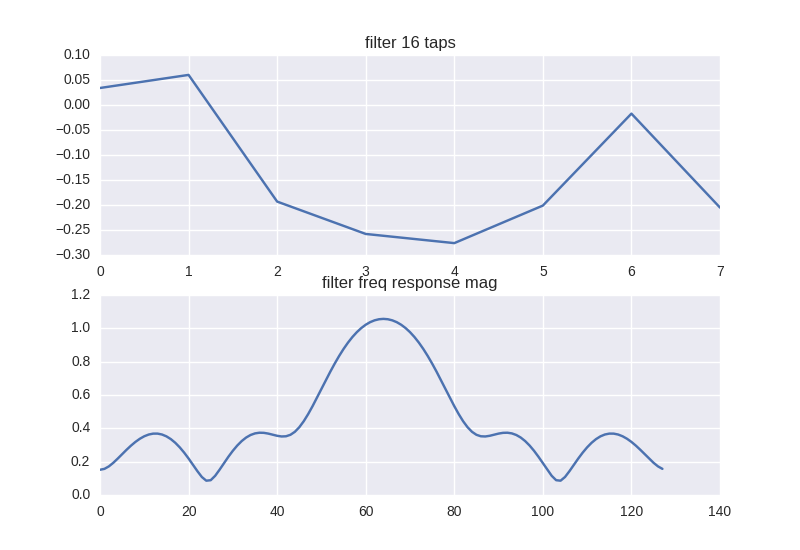}
    \caption{\label{fig:psk-taps}Time and frequency magnitude representations
of a filter in the first convolutional layer of our trained CLDNN.}
  \end{subfigure}
  \begin{subfigure}{\columnwidth}
    \includegraphics[width=\columnwidth]{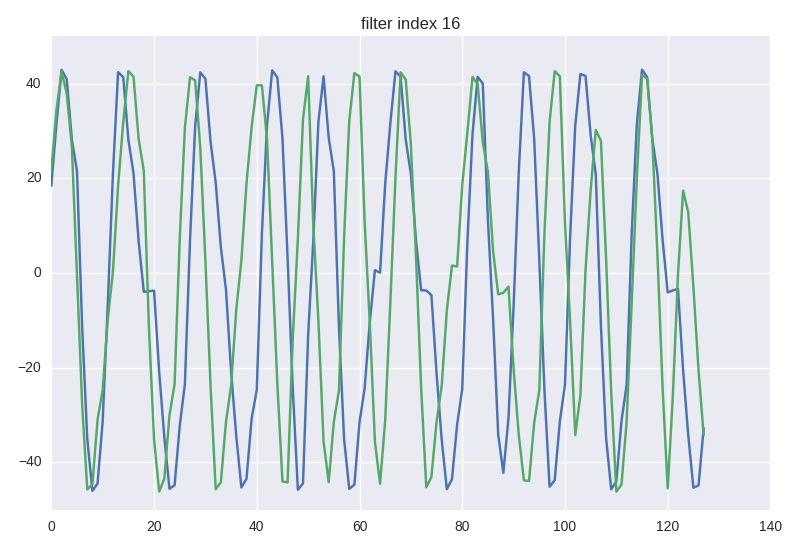}
    \caption{\label{fig:psk-taps-activations}Random data trained to maximally
activate the filter, which winds up looking like BPSK.}
  \end{subfigure}
\end{figure}

\begin{figure}
  \begin{subfigure}{\columnwidth}
    \includegraphics[width=\columnwidth]{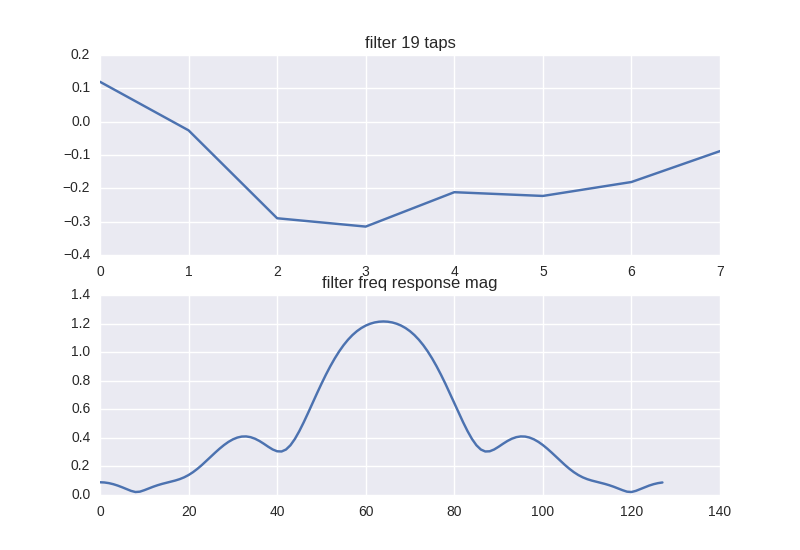}
    \caption{\label{fig:fsk-taps}Time and frequency magnitude representations
of a filter in the first convolutional layer of our trained CLDNN.}
  \end{subfigure}
  \begin{subfigure}{\columnwidth}
    \includegraphics[width=\columnwidth]{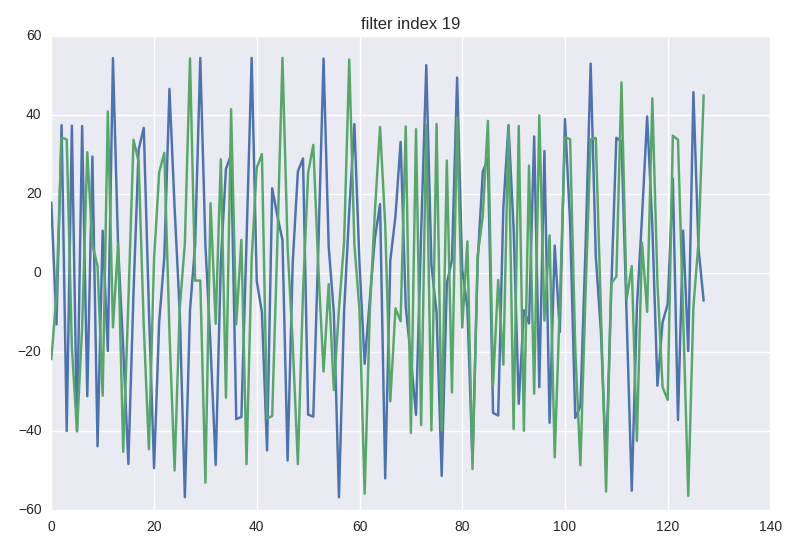}
    \caption{\label{fig:fsk-taps-activations}Random data trained to maximally
activate the filter, which winds up looking like FM or FSK modulation.}
  \end{subfigure}
\end{figure}

Gaining intuition on what the CLDNN is learning in each layer is important for
guiding future work. To do this we plot the time and frequency representations
of some filter taps. For the frequency response the filter taps are zero-padded
with 100 zeros to get a 128-point FFT. Figs. \ref{fig:psk-taps} and
\ref{fig:fsk-taps} show two select filters from the first layer. The
time-domain representations do not look particularly familiar to an expert eye;
however the frequency responses do show shaped low-pass filters. Other filters
that are not shown have frequency selective components, DC blockers, and
sinc-like spectral shapes.

Another way to visualize these filters is to apply random data to them and
perform a gradient ascent for the output of a particular filter which will
converge on data that most activates a convolutional neuron \cite{deepdream}.
Results for the selected two filters are shown in figs.
\ref{fig:psk-taps-activations} and \ref{fig:fsk-taps-activations}. The resulting
vectors look somewhat like crude PSK and FM/FSK modulations to an expert eye.
The vectors also display some constant phase rotation that is present in our
dataset due to the simulated channel model. It is important to note that these
two filter visualizations were selected and not all filters appear meaningful to
an expert.

\section{Discussion}
\label{sec:discussion}

Performance of deep neural networks in the radio domain does not seem to be
limited by network depth the same way the image, natural language processing,
and acoustic domains are. Although our experiments focused on modulation
recognition as a benchmark task, we expect other radio machine learning tasks
to be able to use similar network architectures. Further advances in deep
learning for radio tasks will likely come from improved training methods and
network architectures that can learn to transform RF data to remove effects of
wireless channels, which neural network architectures are not designed for. One
example that is currently being explored is the use of spatial transforms to
equalize and synchronize incoming waveforms \cite{rtn}.

These experiments also focused on a dataset that is nominally
bandwidth-normalized which is a poor assumption for signals captured from real
radio transmissions. Future networks used in real-world applications will need
to learn to either resample signals to be bandwidth normalized, or learn
features for many bandwidths. Networks that can resample, synchronize, and
remove non-linear channel distortions are all exciting future work for the
field.  We believe that as radio environments become increasingly complex,
combining varying temporal behaviors of modulations, multi-modulation protocols
and combining many radio emitters interoperating within a single band, many of
these notions of hierarchy within deep networks will become increasingly important
in allowing our networks to scale to cope with the complexity effectively as
has been similarly shown in the vision domain within complex multi-object scenes.

\bibliography{dl-biblio.bib}
\bibliographystyle{plain}
\end{document}